\documentclass{article}

\usepackage{microtype}
\usepackage{graphicx}
\usepackage{booktabs} 
\usepackage{amsfonts}
\usepackage{amsmath}
\usepackage{amsthm}
\usepackage{caption}
\usepackage{subcaption}

\usepackage{tikz}
\usepackage{pgfplots}
\usepgfplotslibrary{fillbetween}

\newcommand{\arxiv}{}

\DeclareGraphicsExtensions{.png}

\usepackage{hyperref}

\ifdefined\arxiv
\usepackage[accepted,arxiv]{icml2019}
\newcommand{\Appendix}{appendix}
\else
\usepackage{icml2019}
\newcommand{\Appendix}{supplementary materials}
\fi

\newcommand{\om}{label gradient alignment}
\newcommand{\Om}{Label gradient alignment}
\newcommand{\OM}{Label Gradient Alignment}

\newcommand{\NormLMinusU}{\Vert g_\ell - g_u \Vert}

\icmltitlerunning{\OM{}}

\newtheorem{theorem}{Theorem}
\newtheorem{proposition}[theorem]{Proposition}

\title{Semi-Supervised Learning by \OM{}}

\begin{document}

\twocolumn[
\icmltitle{Semi-Supervised Learning by \OM{}}



\icmlsetsymbol{equal}{*}

\begin{icmlauthorlist}
\icmlauthor{Jacob Jackson}{openai,waterloo}
\icmlauthor{John Schulman}{openai}
\end{icmlauthorlist}

\icmlaffiliation{openai}{OpenAI}
\icmlaffiliation{waterloo}{University of Waterloo, Canada}

\icmlcorrespondingauthor{Jacob Jackson}{jbfjacks@uwaterloo.ca}

\icmlkeywords{Machine Learning, Semi-Supervised Learning, Deep Learning, ICML}

\vskip 0.3in
]



\printAffiliationsAndNotice{}  

\begin{abstract}
We present \om{}, a novel algorithm for semi-supervised learning which imputes labels
for the unlabeled data and trains on the imputed labels.
We define a semantically meaningful distance metric on the input space by
mapping a point $(x,y)$ to the gradient of the model at $(x,y)$.
We then formulate an optimization problem whose objective is to minimize the
distance between the labeled and the unlabeled data in this space,
and we solve it by gradient descent on the imputed labels.
We evaluate \om{} using the standardized architecture introduced by \citet{rse} and demonstrate state-of-the-art accuracy in semi-supervised CIFAR-10 classification.
\end{abstract}

\section{Introduction}
In many machine learning applications, obtaining unlabeled data is easy,
but obtaining labeled data is not.
Consider the task of video classification:
an ever-growing number of videos are available for free on the internet,
but obtaining labels for these videos continues to require costly human input.
For this reason, semi-supervised learning (SSL) is an area of growing interest.
SSL seeks to use unlabeled data together with a small amount of labeled data
to produce a better model than could be obtained from the labeled data alone.

Many SSL methods depend on the \textit{semi-supervised smoothness assumption} described in \citet{Chapelle:2010:SL:1841234}:
\begin{center}
\textit{If two points $x_1, x_2$ in a high-density region are close,
then so should be the corresponding outputs $y_1, y_2$.}
\end{center}
With the rise of deep learning, a central question in SSL is how to define ``close'' in a way that
allows the model to exploit its learned representations.
Various methods have been proposed to address this question.
\citet{pmlr-v80-kamnitsas18a} and \citet{Husser2017LearningBA} learn to map the data into a latent space such that the dot product metric
in the latent space is meaningful.
The current state-of-the-art methods for deep semi-supervised image classification
are \textit{consistency regularization} methods.
These methods do not compute distances between points from the input data.
Instead, they synthesize new training inputs $x'$ from the original inputs $x$ such that
$x$ and $x'$ are known to be close in a semantically meaningful space.
Then, they enforce consistency between the classification of $x$ and $x'$.
Methods for obtaining $x'$ include data augmentation, as in the $\Pi$-Model \cite{pimodel},
or adversarial perturbations, as in virtual adversarial training \cite{vat}.

In this work, we seek to directly define a semantically meaningful distance metric
on the input data.
Recognizing the success of gradient similarity in explaining the influence of individual
training points on a deep image classification model \cite{Koh2017UnderstandingBP},
and in identifying helpful auxiliary tasks in reinforcement learning \cite{sim_rl},
we propose to map points into model parameter space $\mathcal{G}$ using the model's gradient:
\begin{equation}
\label{eq:xy_maps}
\varphi(x,y) = \nabla_\theta \, L(\theta, x, y)
\end{equation}
We use the resulting metric to impute labels for the unlabeled points,
similar to the classic technique of label propagation \cite{Zhu02learningfrom}.
However, since $\varphi(x,y)$ depends on $y$, the usual formulation of label propagation does not work.
Instead, we formulate an optimization problem over the labels $y_u$
for the unlabeled data.
The optimization objective is minimization of the Euclidean distance in $\mathcal{G}$
between the labeled data and the unlabeled data.
For common choices of loss function $L(\theta,x,y)$ such as cross entropy and mean squared error,
$\varphi(x,y)$ is linear in $y$, rendering the resulting optimization problem convex.
We solve this optimization problem simultaneously with training the model on the unlabeled data.
We call this technique \textit{\om{}}.

This work is organized as follows.
Section~\ref{sec:method} defines \om{} (LGA).
Section~\ref{sec:motivation} analyzes LGA using simplified settings and synthetic problems.
In Section~\ref{sec:evaluation} we evaluate LGA on standard benchmark datasets.
Finally, we survey related work in Section~\ref{sec:related_work}.

\section{Method}
\label{sec:method}
We describe \om{} in Algorithm~\ref{alg:modified}.
It can be explained in English as follows: we simultaneously impute labels ($y_u$)
for the unlabeled data ($X_u$)
and train the model parameters ($\theta$)
to minimize $L(\theta, X_u, y_u)$ by gradient descent.
The labels $y_u$ are parameterized by $y_u=f(w)$, where $f(x)=x$ (for regression)
or $f(x)=\text{softmax}(x)$ (for classification).
The gradient of the loss on the unlabeled data ($g_u$)
with respect to $\theta$ is a function of $y_u$,
so we can optimize $y_u$ using gradient descent to minimize
the Euclidean distance between $g_u$ and the gradient from the labeled  data ($g_\ell$).

\begin{algorithm}[tb]
   \caption{\OM{}}
   \label{alg:modified}
\begin{algorithmic}
   \STATE {\bfseries Input:}
   \STATE Labeled inputs $X_\ell \in \mathbb{R}^{n_\ell \times m}$
   \STATE Labels $y_\ell \in \mathbb{R}^{n_\ell \times k}$
   \STATE Unlabeled inputs $X_u \in \mathbb{R}^{n_u \times m}$
   \STATE Hyperparameters $\alpha_\theta,\alpha_w\in\mathbb{R}$
   \STATE Initial model parameters $\theta_\text{init} \in \mathbb{R}^p$
   \STATE Loss function $L(\theta, X, y)$
   \STATE Label parameterization function $f$ (softmax for classification, identity function for regression)
   \STATE Labeled gradient coefficient schedule $T(i)$
   \STATE {\bfseries Output:}
   \STATE Learned model parameters $\theta$
   \STATE
   \STATE Initialize $w=\mathbf{0}$ \hspace*{\fill} $w \in \mathbb{R}^{n_u \times k}$
   \STATE Initialize $\theta=\theta_\text{init}$ \hspace*{\fill} $\theta \in \mathbb{R}^p$
   \FOR{$i=1$ {\bfseries to} $N_{\text{iterations}}$}
   \STATE $X_\ell^\text{mini},y_\ell^\text{mini} := \textsc{SampleMinibatch}(X_\ell,y_\ell)$
   \STATE $X_u^\text{mini},w^\text{mini}:= \textsc{SampleMinibatch}(X_u,w)$
   \STATE $y_u^\text{mini} := f(w^\text{mini})$
   \STATE $g_\ell := \nabla_\theta \, L(\theta, X_\ell^\text{mini}, y_\ell^\text{mini})$ \hspace*{\fill} $g_\ell \in \mathbb{R}^{p}$
   \STATE $g_u := \nabla_\theta \, L(\theta, X_u^\text{mini}, y_u^\text{mini})$ \hspace*{\fill} $g_u \in \mathbb{R}^{p}$
   \STATE $g_\theta := g_u + T(i) \, g_\ell$ \hspace*{\fill} $g_\theta \in \mathbb{R}^{p}$
   \STATE $g_w := \nabla_w \Vert \text{EMA}(g_\ell) - g_u \Vert^2_\text{normalized}$ \hspace*{\fill} $g_w \in \mathbb{R}^{n_u \times k}$
   \STATE $\theta \leftarrow \textsc{AdamUpdate}(\alpha_\theta, \theta, g_\theta)$
   \STATE $w \leftarrow \textsc{AdamUpdate}(\alpha_w, w, g_w)$
   \ENDFOR
\end{algorithmic}
\end{algorithm}

We define $\Vert \cdot \Vert^2_\text{normalized}$ as follows,
where $\varepsilon_\text{norm}$ is a hyperparameter introduced for numerical stability:
\begin{equation}
\label{eq:normalize}
\Vert v \Vert^2_\text{normalized} =
\sum_{i=1}^p \frac{v_i^2}{\varepsilon_\text{norm} + \sqrt{\text{EMA}\big(v_i^4\big)} }
\end{equation}
With this normalization, the metric $\Vert \text{EMA}(g_\ell) - g_u \Vert^2_\text{normalized}$
is invariant to the scaling of each parameter $\theta_k$,
assuming $\varepsilon_\text{norm}$ does not have a significant effect.

In Algorithm~\ref{alg:modified} and Equation~\ref{eq:normalize}, we use $\text{EMA}(x)$ to denote an exponential moving
average of the value $x$.
We use exponential moving averages to decrease the variance of our estimate of $x$
when $x$ depends on the sampled minibatch.
We treat the exponential moving average as a constant when computing gradients.

\section{Analysis}
\label{sec:motivation}
\subsection{Linear Regression}
We motivate \om{} by studying it in the simplest possible setting:
a linear regression model.
We make several simplifying modifications to Algorithm~\ref{alg:modified}
so that it is easier to analyze theoretically:
\begin{itemize}
\item We replace Adam \cite{DBLP:journals/corr/KingmaB14} with gradient descent.
\item We do not use minibatches
(that is, we optimize over the whole dataset on each iteration).
\item Since we do not use minibatches, there is no stochasticity in our estimates.
So we replace $\text{EMA}(x)$ with $x$ in every place where $\text{EMA}(x)$ occurs.
(However, we still treat $x$ as a constant when computing gradients.)
\item We ignore the effect of $\varepsilon_\text{norm}$.
\end{itemize}
We will also make a strong assumption on the training data, namely that the
matrices $X_\ell^\top X_\ell$ and $X_u^\top X_u$ are diagonal
(where $X_\ell$ and $X_u$ represent the labeled and unlabeled input data, respectively).
This is equivalent to the assumption that the input features are uncorrelated,
and it is necessary because the normalization in Equation~\ref{eq:normalize} is
affected by a change of basis.

Subject to these simplifying assumptions, 
we will show that \om{} is a regularizer causing the algorithm
to favor models aligned with the principal components of the unlabeled data.
This reveals a connection with other regularizers, such as early stopping,
which favors the principal components of the \textit{labeled} data,
and the truncated singular value decomposition \cite{hansen1987truncatedsvd},
which projects the input data onto
the first $k$ principal components (where $k$ is a hyperparameter).

First, we introduce notation.
Suppose we have labeled training data $X \in \mathbb{R}^{n\times m},
y \in\mathbb{R}^n$, and unlabeled training data $X_u\in\mathbb{R}^{n_u \times m}$.
The linear regression model, parameterized by $\theta \in \mathbb{R}^m$, predicts that $y=\theta^\top x$ for $x\in\mathbb{R}^m$.
The training loss $L(\theta)$ is given by:
\begin{equation}
L(\theta) = \frac{1}{2n} \Vert y-X\theta \Vert^2
\end{equation}

Initializing $\theta_0=0$ and applying gradient descent with learning rate $\alpha \in \mathbb{R}$, we obtain iterates $\theta_0, \theta_1, \dots$ defined by the recurrence:
\begin{equation}
\theta_{k+1} = \theta_k - \alpha \, \nabla L(\theta_k)
\end{equation}

The minimizer of $L(\theta)$ is:
\begin{equation}
\label{eq:lse}
\theta^* = (X^\top X)^{-1}X^\top y
\end{equation}
Let $\lambda_1,\dots,\lambda_m$ and $q_1,\dots,q_m$ be eigenvalues and orthogonal unit eigenvectors of $\frac{1}{n} X^\top X$.
From Equation~\ref{eq:lse} we obtain an alternate formula for $\theta^*$:
\begin{equation}
\label{eq:eigsum}
    \theta^* = \frac{1}{n} \sum_{i=1}^m \frac{1}{\lambda_i} q_i q_i^\top X^\top y
\end{equation}

The iterates $\theta_k$ admit the following closed form \cite{goh2017why}:
\begin{equation}
\label{eq:gdclosed}
    \theta_k = \frac{1}{n}\sum_{i=1}^m \frac{1-(1-\alpha \lambda_i)^k}{\lambda_i} \, q_i q_i^\top X^\top y
\end{equation}
We would like to understand the effect of \om{} when applied in the linear regression setting.
As we will see, the fixed points of Algorithm~\ref{alg:modified} are the same as the fixed points
of gradient descent on $L(\theta,X,y)$, that is,
\om{} has the same fixed points as the fully supervised model.
For this reason, we study the effect of \om{} by considering the dynamics during training.

There is a well-known technique which also depends on the dynamics during training
rather than the fixed point at convergence: early stopping.
Early stopping is a regularization technique which uses $\theta_k$ as the trained model,
where $k$ selected by some rule (typically, to minimize validation error).
The term $1-(1-\alpha \lambda_i)^k$ in Equation~\ref{eq:gdclosed} converges to $1$ quickly when $\lambda_i$ is large.
Thus, Equation~\ref{eq:gdclosed} shows that early stopping is a spectral regularizer which increases the relative influence of the terms in Equation~\ref{eq:eigsum} corresponding to the largest eigenvalues of $X^\top X$, also known as the principal components of $X$.

The principle underlying early stopping can be stated succinctly as follows:
\textit{models which are learned quickly generalize better.}
From Equation~\ref{eq:gdclosed}, we see that this is equivalent to the following principle:
\textit{models aligned with the principal components of the labeled data generalize better.}
The principal components are determined by $X$: they do not depend on $y$.
Thus, to the extent that the previous principle holds, the following principle also holds:
\textit{models aligned with the principal components of the unlabeled data generalize better.}
The effect of \om{}, then, is to cause the model to learn faster in the direction of the
principal components of the unlabeled data.
When combined with early stopping, this causes the selected model to favor these features.

\begin{figure*}
\centering
\begin{subfigure}[b]{.49\textwidth}
\centering
\includegraphics{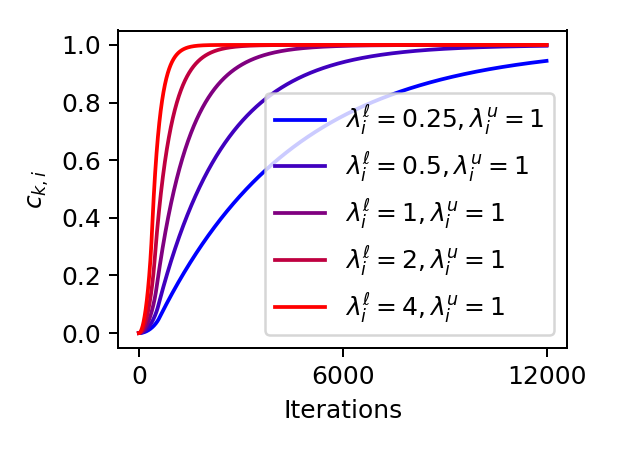}
\end{subfigure}
\begin{subfigure}[b]{.49\textwidth}
\centering
\includegraphics{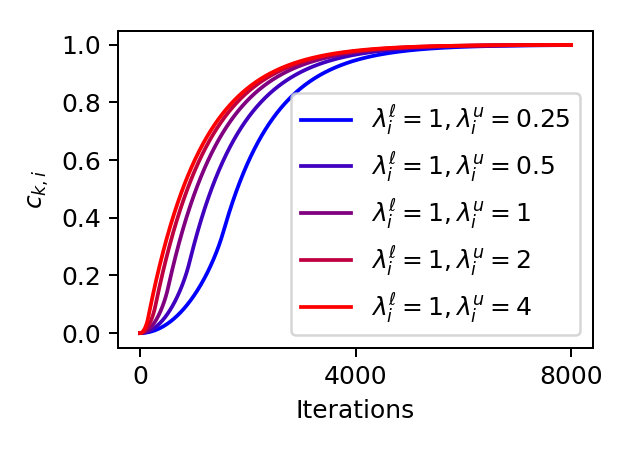}
\end{subfigure}
\caption{
Values of $c_{k,i}$ during training for various values of
$\lambda^\ell_i$ and $\lambda^u_i$.
The red lines converge to $1$ faster,
showing that increasing either $\lambda^\ell_i$ or $\lambda^u_i$
causes the model to learn faster in the correponding direction $e_i$.
The hyperparameters were $\alpha_\theta=\alpha_w=\varepsilon_\text{norm}=10^{-3}$.
$\frac{1}{n_\ell} X^\top_\ell y_\ell$ was kept constant within each plot.
}
\label{fig:learnspeeds}
\end{figure*}

We will show that \om{} causes the model to learn the terms in Equation~\ref{eq:eigsum} corresponding to the principal components of $X_u$ more quickly than the other eigenvectors of $X_u^\top X_u$.
To permit our simplified analysis, we assume that $X^\top X$ and $X^\top_u X_u$
are diagonal.\footnote{This assumption is necessary because
the normalization procedure defined in Equation~\ref{eq:normalize}
is affected by a change of basis.}
In this case, the eigenvectors $q_1,\dots,q_m$ are the unit vectors $e_1,\dots,e_m$.
For clarity we add $\ell$ subscripts or superscripts to $X$, $y$, and $\lambda_i$
to indicate that they correspond to the labeled data.
We denote the eigenvalues of $\frac{1}{n_u} X^\top_u X_u$ as $\lambda^u_1,\dots,\lambda^u_m$
(by the assumption that $X^\top_u X_u$ is diagonal,
we have $\lambda^u_i=\frac{1}{n_u} e_ie_i^\top X^\top_u X_u$).
We introduce a learned parameter vector $y^u \in \mathbb{R}^{n_u}$
representing the imputed labels for $X_u$.
The update rule (from Algorithm~\ref{alg:modified}) is:
\begin{align}
g^\ell_k &= \nabla_{\theta_k} \frac{1}{2n_\ell} \Vert y_\ell - X_\ell \theta_k \Vert^2 \\
g^u_k &= \nabla_{\theta_k} \frac{1}{2n_u} \Vert y_u - X_u \theta_k \Vert^2 \\
\theta_{k+1} &= \theta_k - \alpha_\theta g^u_k \\
y^u_{k+1} &= y^u_k - \alpha_w \nabla_{y^u_k} \frac{1}{2} \Vert g^\ell_k - g^u_k \Vert^2_\text{normalized}
\end{align}
Define $c_k \in \mathbb{R}^m$ such that:
\begin{equation}
\theta_k = \frac{1}{n_\ell} \sum_{i=1}^m \frac{c_{k,i}}{\lambda_i^\ell} e_i e_i^\top X^\top_\ell y_\ell
\end{equation}
We desire that $c_{k,i}$ should converge to $1$ faster when $\lambda^u_i$ is large.
\begin{proposition}
\label{prop:no_dependence}
Let $i \ne j$. Then the value of $c_{k,i}$ does not depend on
$\lambda^\ell_j$ or $\lambda^u_j$.
(The proof is in the \Appendix{}.)
\end{proposition}
Proposition~\ref{prop:no_dependence} shows that in this setting,
we can understand the effect of \om{} by considering each dimension separately.
Thus, we study the effect of \om{} on $c_{k,i}$ by varying
$\lambda_i^u$ and $\lambda_i^\ell$.
Figure~\ref{fig:learnspeeds} plots $c_{k,i}$ for various values of
$\lambda_i^u$ and $\lambda_i^\ell$.
Empirically, we observe faster convergence for larger $\lambda^u_i$, as desired.

We also see in Figure~\ref{fig:learnspeeds}
that each $c_{k,i}$ converges to $1$, that is,
\om{} converges to the same fixed point as fully supervised gradient descent.
As the following propositions will show,
the fixed points of \om{} and the fully supervised algorithm
are always the same when the loss function is convex.
Thus, early stopping is essential when using \om{} with
convex loss functions such as in linear regression.
\begin{proposition}
\label{prop:fixpoint}
Let $(\theta^*, w^*)$ be a fixed point of Algorithm~\ref{alg:modified},
and suppose that $\NormLMinusU^2=0$ is attainable for some $w$.
Let $\nabla_\theta \, L(\theta, X_u, y_u)$ be linear in $y_u$.\footnote{
This is satisfied for common loss functions such as cross entropy and mean squared error.}
Then $\nabla_{\theta^*} \, L(\theta^*, X_\ell, y_\ell) = 0$.
\end{proposition}
\begin{proof}
Since $\NormLMinusU^2$ is convex in $y_u$,
and we assumed that $\NormLMinusU^2=0$ is attainable,
the fact that $y_u^*$ is a fixed point implies that $g_\ell = g_u$.
Then the fact that $\theta^*$ is a fixed point implies $g_u = 0$.
\end{proof}
The following proposition is a corollary of Proposition~\ref{prop:fixpoint}.
\begin{proposition}
Let $L(\theta, X_\ell, y_\ell)$ be convex in $\theta$.
Then if $(\theta^*, y^*)$ is a fixed point of Algorithm~\ref{alg:modified},
$\theta^*$ is the global minimum of $L(\theta, X_\ell, y_\ell)$.
That is, the unique fixed point is the same as that of the fully supervised algorithm.
\end{proposition}
This proposition is why throughout Section~\ref{sec:motivation},
we seek to understand \om{} by its intermediate states and its updates, rather than the model parameters at convergence.

\subsection{A Nonlinear Example}
In the previous subsection, we gave an explanation for why \om{} works in
simplified setting:
it uses the unlabeled data to estimate the principal components.
One naturally wonders: to what extent does this intuition apply to the nonlinear setting?
In this subsection, we seek to answer this question through experiments on a synthetic dataset.

\begin{figure}[t]
\begin{center}
\includegraphics[scale=0.6]{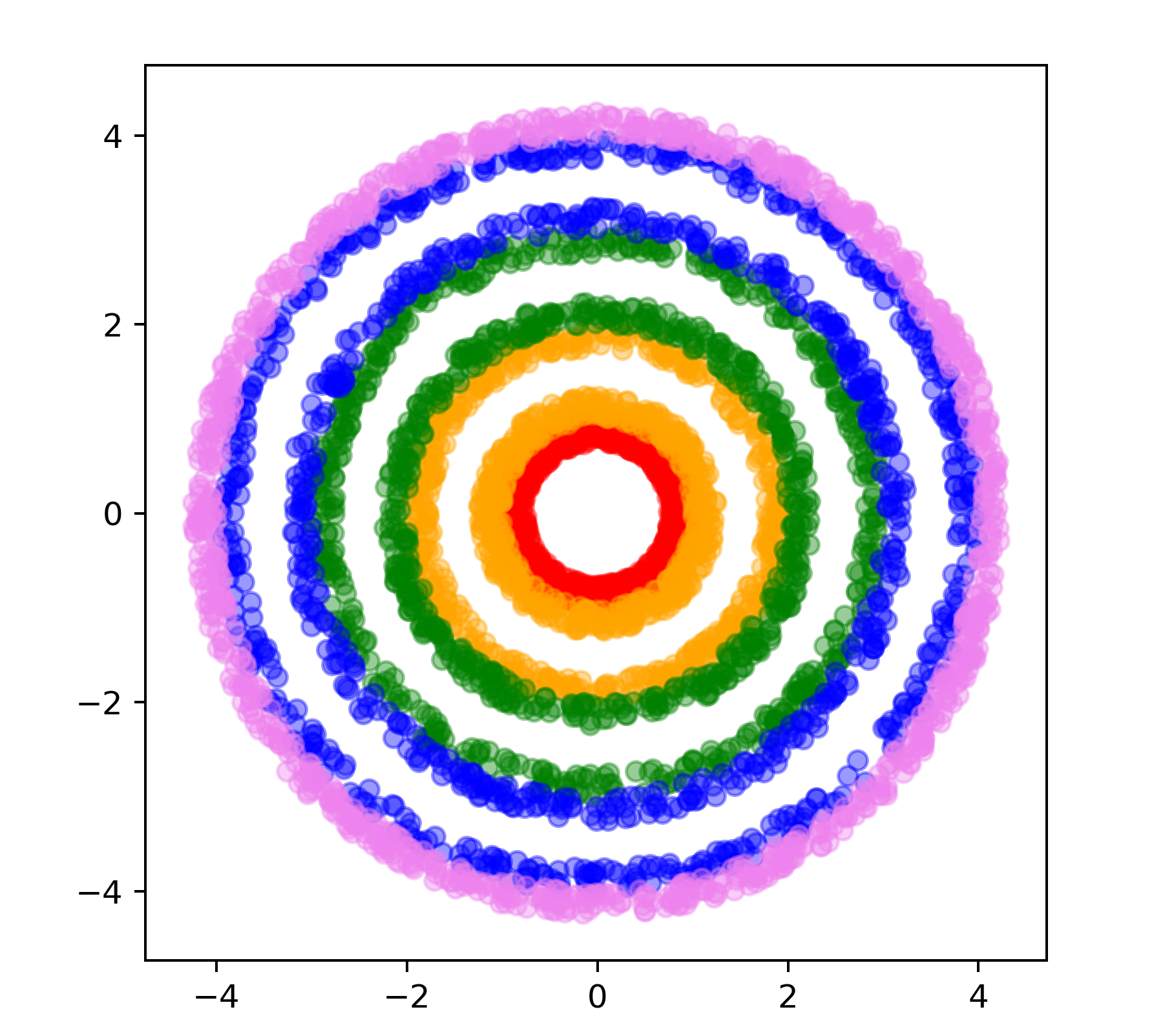}
\end{center}
\caption{The synthetic dataset with $d=2$ and $n=2500$. Experiments use $d=50$ and $n=5000$.
The color of a point indicates its class label.}
\label{fig:synth_picture}
\end{figure}

First, we define the metric that we will measure.
We note that in the linear regression setting,
the principal components are the same as the principal eigenvectors of the Hessian $\nabla^2_\theta L(\theta) = X^\top X$.
Motivated by this fact, we define the \textit{alignment} of an update to be its cosine similarity
with the principal eigenvector of the Hessian $\nabla^2 L(\theta, X_t, y_t)$,
where $X_t$ and $y_t$ denote the test inputs and labels respectively.
That is, if $\nabla_\theta $ is the update to the model parameters $\theta$
and $q_1$ is the principal eigenvector of $\nabla^2 L(\theta, x_t, y_t)$, the alignment is:
\begin{equation}
\label{eq:alignment}
\text{alignment}(\nabla_\theta) = \frac{| q_1^\top \nabla_\theta |}{ \Vert q_1 \Vert\Vert \nabla_\theta\Vert }
\end{equation}
We will see that \om{} increases the alignment of the updates.

\begin{figure}
\begin{tikzpicture} \begin{axis}[xlabel=Iterations,
ylabel=Alignment,
ymin=0,
ymax=1,
legend pos=north west]
\addplot[color=red, mark=triangle] coordinates { (0, 0.5503208281099796) (25, 0.22984331905841826) (50, 0.33189630299806594) (75, 0.2587306619435549) (100, 0.4210384462028742) (125, 0.45332514453679323) (150, 0.5783295797556639) (175, 0.6611670001037419) (200, 0.663612950667739) (225, 0.7284115254878998) (250, 0.8293520468473434) (275, 0.9073228740692139) (300, 0.8856144171953201) (325, 0.8481157076358795) (350, 0.9130972445011138) (375, 0.9193143963813781) (400, 0.8968876168131829) (425, 0.9332962369918824) (450, 0.9314273478090763) (475, 0.9314598661661148) (500, 0.9607370710372924) (525, 0.9461282968521119) (550, 0.897813755273819) (575, 0.9208642709255218) };
\addlegendentry{ LGA }
\addplot[color=blue, mark=diamond] coordinates { (0, 0.4276863223314285) (25, 0.17724458277225494) (50, 0.1026847992092371) (75, 0.1390729020535946) (100, 0.15557827942073346) (125, 0.15074963495135307) (150, 0.14444458853453399) (175, 0.14020133800804616) (200, 0.13986071515828372) (225, 0.1462627040594816) (250, 0.16618605084717275) (275, 0.185114294141531) (300, 0.1974946340546012) (325, 0.2022970824688673) (350, 0.20594847232103347) (375, 0.20195110708475114) (400, 0.19006424373015762) (425, 0.18028906237334014) (450, 0.17309703275561333) (475, 0.1658038341253996) (500, 0.1609668789803982) (525, 0.15944130659103395) (550, 0.15710300743579864) (575, 0.15457946076989174) };
\addlegendentry{ Supervised }
\addplot[name path=upper,draw=none] coordinates { (0, 0.854764597481009) (25, 0.40978486069283143) (50, 0.5113583248209617) (75, 0.43791268445538023) (100, 0.7168135417713659) (125, 0.7463994903930101) (150, 0.8836488961131348) (175, 0.9641925576374704) (200, 1.0027034089775642) (225, 0.998275924655959) (250, 1.0559200185879978) (275, 1.02720991423847) (300, 1.1058212509685788) (325, 1.0653552608618606) (350, 1.0806349948414207) (375, 1.0344422925665655) (400, 1.118020038127042) (425, 1.0347342782970133) (450, 1.1205055374835258) (475, 1.1070438236268805) (500, 1.0497282344538057) (525, 1.0819323847490154) (550, 1.0624647687129296) (575, 1.0532563762302072) };
\addplot[name path=lower,draw=none] coordinates { (0, 0.24587705873895022) (25, 0.04990177742400509) (50, 0.15243428117517016) (75, 0.07954863943172952) (100, 0.12526335063438243) (125, 0.16025079868057635) (150, 0.27301026339819306) (175, 0.3581414425700134) (200, 0.3245224923579138) (225, 0.4585471263198405) (250, 0.6027840751066892) (275, 0.7874358338999576) (300, 0.6654075834220614) (325, 0.6308761544098984) (350, 0.7455594941608071) (375, 0.8041865001961908) (400, 0.6757551954993238) (425, 0.8318581956867516) (450, 0.742349158134627) (475, 0.7558759087053492) (500, 0.8717459076207793) (525, 0.8103242089552084) (550, 0.7331627418347084) (575, 0.7884721656208364) };
\addplot[fill=red!10] fill between[of=upper and lower];
\addplot[name path=upper,draw=none] coordinates { (0, 0.6674330425385887) (25, 0.30561497368996204) (50, 0.1697717498455109) (75, 0.20760792256819982) (100, 0.2276727674855361) (125, 0.2160131133471231) (150, 0.20765676120558105) (175, 0.19751914366126483) (200, 0.20531778735279732) (225, 0.21799651584394783) (250, 0.2467373933165784) (275, 0.2661621347324296) (300, 0.28082687458421296) (325, 0.27888889302259035) (350, 0.2822276452501725) (375, 0.2738944338791395) (400, 0.25858972009721687) (425, 0.24267555488541312) (450, 0.23023955742306546) (475, 0.2194243159286753) (500, 0.21167644847254435) (525, 0.2089477183920616) (550, 0.20226528326494925) (575, 0.19905676552267826) };
\addplot[name path=lower,draw=none] coordinates { (0, 0.18793960212426827) (25, 0.048874191854547805) (50, 0.03559784857296329) (75, 0.07053788153898936) (100, 0.08348379135593083) (125, 0.08548615655558307) (150, 0.08123241586348692) (175, 0.0828835323548275) (200, 0.07440364296377011) (225, 0.07452889227501538) (250, 0.08563470837776711) (275, 0.10406645355063238) (300, 0.1141623935249894) (325, 0.12570527191514425) (350, 0.12966929939189442) (375, 0.1300077802903628) (400, 0.12153876736309836) (425, 0.11790256986126715) (450, 0.11595450808816121) (475, 0.11218335232212387) (500, 0.11025730948825202) (525, 0.10993489479000629) (550, 0.11194073160664801) (575, 0.11010215601710523) };
\addplot[fill=blue!10] fill between[of=upper and lower];

\end{axis}
\end{tikzpicture}
\caption{
Alignment (defined in Equation~\ref{eq:alignment}) of the parameter updates of each algorithm over the course of training.
Shaded regions indicate standard deviation over 25 trials.}
\label{fig:synth_alignment}
\end{figure}
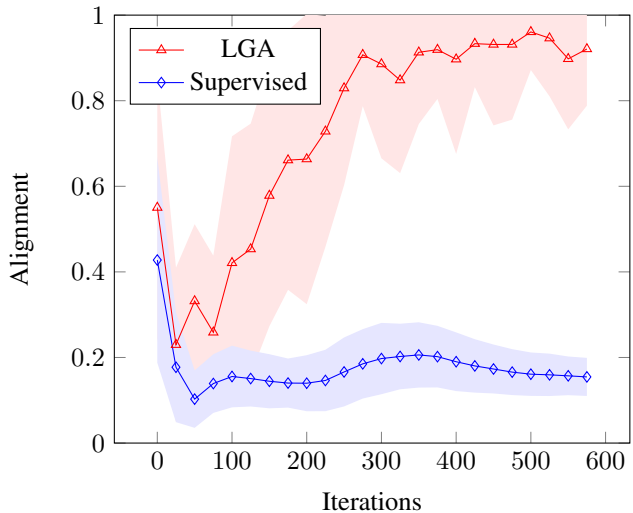

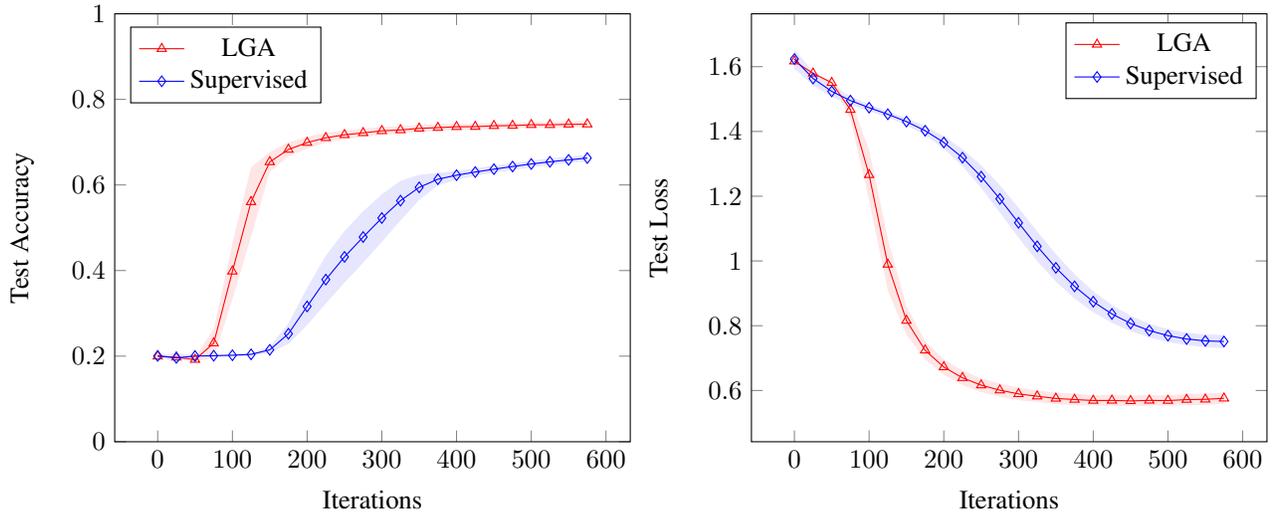
\begin{figure*}
\centering
\begin{subfigure}[b]{.49\textwidth}
\centering
\begin{tikzpicture} \begin{axis}[xlabel=Iterations,
ylabel=Test Accuracy,
ymin=0,
ymax=1,
legend pos=north west]
\addplot[color=red, mark=triangle] coordinates { (0, 0.19962560057640075) (25, 0.1974959999322891) (50, 0.19188319981098176) (75, 0.2304335981607437) (100, 0.3982207977771759) (125, 0.5604975986480712) (150, 0.6537839984893798) (175, 0.6829407978057861) (200, 0.6990176033973694) (225, 0.7101919960975647) (250, 0.7171423959732056) (275, 0.7217151975631714) (300, 0.7264591956138611) (325, 0.7284880042076111) (350, 0.7322143983840942) (375, 0.7339680051803589) (400, 0.7359775948524475) (425, 0.7362544059753418) (450, 0.7383119916915893) (475, 0.7388159990310669) (500, 0.740619204044342) (525, 0.7405663990974426) (550, 0.7417504048347473) (575, 0.7419792008399964) };
\addlegendentry{ LGA }
\addplot[color=blue, mark=diamond] coordinates { (0, 0.20061600029468538) (25, 0.19606240093708038) (50, 0.20019840121269225) (75, 0.20099839806556702) (100, 0.20209280014038086) (125, 0.20416160106658934) (150, 0.21477760136127472) (175, 0.25239839911460876) (200, 0.3161392003297806) (225, 0.3788720011711121) (250, 0.4319327986240387) (275, 0.47836960315704347) (300, 0.5227680015563965) (325, 0.5633568024635315) (350, 0.5944752097129822) (375, 0.6134080052375793) (400, 0.6228528046607971) (425, 0.6301615977287293) (450, 0.6368528056144714) (475, 0.6431631994247436) (500, 0.6490432000160218) (525, 0.6541424036026001) (550, 0.6586495971679688) (575, 0.6628719997406006) };
\addlegendentry{ Supervised }
\addplot[name path=upper,draw=none] coordinates { (0, 0.20122459465692094) (25, 0.200682632087892) (50, 0.19758595776112797) (75, 0.2637807378879314) (100, 0.4639336407723432) (125, 0.6416470880094486) (150, 0.6769985310265858) (175, 0.6994581378112251) (200, 0.7127375330010381) (225, 0.7219404898596021) (250, 0.7274102458638977) (275, 0.7315413297883833) (300, 0.7357850772485239) (325, 0.736842693987445) (350, 0.7403171716715113) (375, 0.7422598438452912) (400, 0.7432522498060118) (425, 0.7439054163468515) (450, 0.7446229442130297) (475, 0.7453927945541805) (500, 0.7469114295331387) (525, 0.7476873636719703) (550, 0.7484937544607044) (575, 0.7490516236994565) };
\addplot[name path=lower,draw=none] coordinates { (0, 0.19802660649588055) (25, 0.19430936777668623) (50, 0.18618044186083554) (75, 0.19708645843355604) (100, 0.33250795478200856) (125, 0.47934810928669386) (150, 0.6305694659521739) (175, 0.6664234578003471) (200, 0.6852976737937007) (225, 0.6984435023355272) (250, 0.7068745460825135) (275, 0.7118890653379595) (300, 0.7171333139791982) (325, 0.7201333144277772) (350, 0.724111625096677) (375, 0.7256761665154265) (400, 0.7287029398988832) (425, 0.728603395603832) (450, 0.732001039170149) (475, 0.7322392035079532) (500, 0.7343269785555453) (525, 0.7334454345229149) (550, 0.7350070552087902) (575, 0.7349067779805362) };
\addplot[fill=red!10] fill between[of=upper and lower];
\addplot[name path=upper,draw=none] coordinates { (0, 0.2023746910558111) (25, 0.2066431085673791) (50, 0.20236480508878174) (75, 0.2034365869792234) (100, 0.2044951110567447) (125, 0.20692115902214925) (150, 0.22245069883889038) (175, 0.27624380997719933) (200, 0.35958729953012447) (225, 0.43472811722663224) (250, 0.4921986194079841) (275, 0.538414863221222) (300, 0.5781512029489846) (325, 0.609780849284778) (350, 0.6238026390611869) (375, 0.6279449961276887) (400, 0.6332383665528011) (425, 0.6388801902398226) (450, 0.6450005772813662) (475, 0.6511800999743821) (500, 0.6567407266989502) (525, 0.6618722237777008) (550, 0.6661501813355389) (575, 0.6702597216072493) };
\addplot[name path=lower,draw=none] coordinates { (0, 0.19885730953355965) (25, 0.18548169330678166) (50, 0.19803199733660276) (75, 0.19856020915191064) (100, 0.19969048922401703) (125, 0.20140204311102944) (150, 0.20710450388365906) (175, 0.22855298825201822) (200, 0.27269110112943673) (225, 0.3230158851155919) (250, 0.37166697784009334) (275, 0.418324343092865) (300, 0.46738480016380834) (325, 0.5169327556422849) (350, 0.5651477803647774) (375, 0.5988710143474699) (400, 0.612467242768793) (425, 0.621443005217636) (450, 0.6287050339475767) (475, 0.6351462988751052) (500, 0.6413456733330933) (525, 0.6464125834274994) (550, 0.6511490130003986) (575, 0.6554842778739519) };
\addplot[fill=blue!10] fill between[of=upper and lower];

\end{axis}
\end{tikzpicture}
\end{subfigure}
\begin{subfigure}[b]{.49\textwidth}
\centering
\begin{tikzpicture} \begin{axis}[xlabel=Iterations,
ylabel=Test Loss]
\addplot[color=red, mark=triangle] coordinates { (0, 1.61683997631073) (25, 1.57938316822052) (50, 1.5504324960708618) (75, 1.4672226428985595) (100, 1.2660106992721558) (125, 0.9893159818649292) (150, 0.8158691453933716) (175, 0.7240736722946167) (200, 0.6723110270500183) (225, 0.6389807391166688) (250, 0.6163708019256592) (275, 0.6008445882797241) (300, 0.5893462133407593) (325, 0.582163724899292) (350, 0.5753019714355468) (375, 0.5720102548599243) (400, 0.5689424800872803) (425, 0.5691641306877137) (450, 0.5677577519416809) (475, 0.5692454504966736) (500, 0.5687123584747314) (525, 0.571807279586792) (550, 0.5726287221908569) (575, 0.5757119870185852) };
\addlegendentry{ LGA }
\addplot[color=blue, mark=diamond] coordinates { (0, 1.6237395191192627) (25, 1.563472762107849) (50, 1.5236739063262938) (75, 1.495302038192749) (100, 1.4733905649185182) (125, 1.4530593299865722) (150, 1.430376214981079) (175, 1.402223415374756) (200, 1.365939998626709) (225, 1.3190385341644286) (250, 1.2604161834716796) (275, 1.191692485809326) (300, 1.1178666687011718) (325, 1.0451067781448364) (350, 0.9787035346031189) (375, 0.9214498496055603) (400, 0.8741184639930725) (425, 0.8363001799583435) (450, 0.8069222378730774) (475, 0.7849816465377808) (500, 0.7692922449111939) (525, 0.7589067745208741) (550, 0.7530634546279907) (575, 0.7512435150146485) };
\addlegendentry{ Supervised }
\addplot[name path=upper,draw=none] coordinates { (0, 1.643646366153864) (25, 1.5813705667488647) (50, 1.5579281140735466) (75, 1.4966570253407383) (100, 1.3448071399676826) (125, 1.070356613186935) (150, 0.8600102484401747) (175, 0.7519786033482109) (200, 0.6960604612808934) (225, 0.6614268963460433) (250, 0.637673789484987) (275, 0.6211551624703224) (300, 0.6088325874551079) (325, 0.600078778747168) (350, 0.5930496894339751) (375, 0.5899401665024445) (400, 0.5849909418218673) (425, 0.5862351157280967) (450, 0.5837607444589621) (475, 0.5853956801236703) (500, 0.5847702367642763) (525, 0.5891099090211119) (550, 0.5898502979267273) (575, 0.5932618495757844) };
\addplot[name path=lower,draw=none] coordinates { (0, 1.5900335864675958) (25, 1.5773957696921752) (50, 1.542936878068177) (75, 1.4377882604563808) (100, 1.187214258576629) (125, 0.9082753505429233) (150, 0.7717280423465686) (175, 0.6961687412410226) (200, 0.6485615928191433) (225, 0.6165345818872942) (250, 0.5950678143663314) (275, 0.5805340140891257) (300, 0.5698598392264107) (325, 0.5642486710514161) (350, 0.5575542534371186) (375, 0.5540803432174041) (400, 0.5528940183526934) (425, 0.5520931456473306) (450, 0.5517547594243997) (475, 0.553095220869677) (500, 0.5526544801851866) (525, 0.5545046501524721) (550, 0.5554071464549866) (575, 0.5581621244613859) };
\addplot[fill=red!10] fill between[of=upper and lower];
\addplot[name path=upper,draw=none] coordinates { (0, 1.6531977152033424) (25, 1.58466809296639) (50, 1.5410300288245395) (75, 1.5085671174738482) (100, 1.4836355535518806) (125, 1.4624409699799419) (150, 1.4408849039667389) (175, 1.415843757725602) (200, 1.3848095519161356) (225, 1.3452687328283066) (250, 1.2951142144678947) (275, 1.2338463681139162) (300, 1.1640575122395993) (325, 1.0910193033800248) (350, 1.0211596129218115) (375, 0.9588880461291034) (400, 0.9062789552484737) (425, 0.8637897523058768) (450, 0.8306529596283556) (475, 0.8060160754949997) (500, 0.7888706748979191) (525, 0.778081226130911) (550, 0.772548335176039) (575, 0.7717029767927936) };
\addplot[name path=lower,draw=none] coordinates { (0, 1.594281323035183) (25, 1.542277431249308) (50, 1.5063177838280482) (75, 1.48203695891165) (100, 1.4631455762851557) (125, 1.4436776899932025) (150, 1.4198675259954192) (175, 1.3886030730239098) (200, 1.3470704453372822) (225, 1.2928083355005506) (250, 1.2257181524754646) (275, 1.149538603504736) (300, 1.0716758251627443) (325, 0.999194252909648) (350, 0.9362474562844264) (375, 0.8840116530820172) (400, 0.8419579727376713) (425, 0.8088106076108101) (450, 0.7831915161177991) (475, 0.7639472175805619) (500, 0.7497138149244686) (525, 0.7397323229108371) (550, 0.7335785740799424) (575, 0.7307840532365033) };
\addplot[fill=blue!10] fill between[of=upper and lower];

\end{axis}
\end{tikzpicture}
\end{subfigure}
\begin{center}
\caption{Test accuracy and loss during training on the synthetic dataset.
Shaded regions indicate standard deviation over 25 trials.}
\label{fig:synth_loss_acc}
\end{center}
\end{figure*}

The experimental setup was as follows.
The dataset was parametrized by $d$, the dimension, and $n$, the number of labeled points.
It is pictured in Figure~\ref{fig:synth_picture} for $d=2$.
We used $d=50$ and $n=5000$.
The number of unlabeled points was $5n = 25\,000$.
Each point $x$ was generated by sampling a vector $v\in\mathbb{R}^d$ from the unit hypersphere
and a class label $c \sim \text{Discrete}(1,\dots,5)$.
The magnitude $s$ was generated by:
\begin{equation}
s \sim c - 1 + \begin{cases}
\text{Unif}[0.75, 1] & \text{$c=1$} \\
\text{Unif}[0, 0.25] & \text{$c=5$} \\
\text{Unif}\big([0, 0.25] \cup [0.75, 1]\big) & \text{$c \in \{2,3,4\}$}
\end{cases}
\end{equation}
The point $x$ was then defined as $x=vs$.
This dataset was constructed so that the optimal decision boundaries
would pass through regions of high density in the input space.
Many semi-supervised learning methods assume that the decision boundary should avoid such regions;
see the \textit{semi-supervised smoothness assumption} described in \citet{Chapelle:2010:SL:1841234}.
We demonstrate that \om{} improves upon the supervised baseline even in this setting.

The model was a 4-layer fully-connected neural network with 128 hidden units and ReLU activations
\cite{Nair:2010:RLU:3104322.3104425}.
We used Algorithm~\ref{alg:modified} for the line marked ``LGA''
and Adam \cite{DBLP:journals/corr/KingmaB14} to train the supervised model.
Both algorithms were trained on $n=5000$ labeled points,
and \om{} was given $5n=25\,000$ additional unlabeled points.

We compare to the supervised baseline in Figure~\ref{fig:synth_loss_acc},
observing an improvement in test loss and accuracy by using \om{}.
Figure~\ref{fig:synth_alignment} shows the alignment (defined in Equation~\ref{eq:alignment})
during training.
We observe that the alignment is larger when using \om{} compared to the supervised baseline.
Since models aligned with the principal components generalize better in the linear regression setting,
we conjecture that this increased alignment is responsible for the improved generalization performance.

\section{Evaluation}
\label{sec:evaluation}
\citet{rse} evaluated several semi-supervised learning (SSL) algorithms using the same model architecture and experimental procedure.
We used their open source code to evaluate \om{} under the same conditions.
Specifically, we used ``WRN-28-2'' \cite{BMVC2016_87},
i.e. ResNet \cite{He2016DeepRL} with depth 28 and width 2, including batch normalization
\cite{Ioffe2015BatchNA} and leaky ReLU nonlinearities,
trained with the Adam optimizer \cite{DBLP:journals/corr/KingmaB14}.
We used the same hyperparameter tuning procedure as \citet{rse} by tuning our method's
hyperparameters separately for CIFAR-10 with 4000 labels and SVHN  with 1000 labels, then using the same hyperparameters for all experiments involving a given dataset.

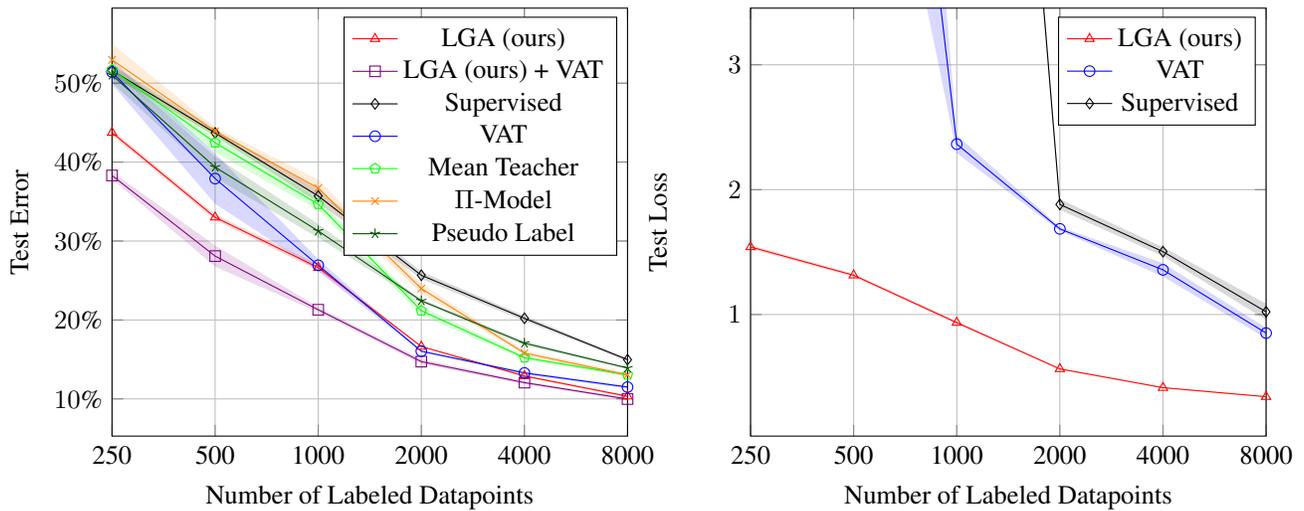
\begin{figure*}
\centering
\begin{subfigure}[b]{.49\textwidth}
\centering
\begin{tikzpicture}
\begin{axis}[
xmode=log,
xlabel=Number of Labeled Datapoints,
grid=major,
xtick=data,
xticklabels={250,500,1000,2000,4000,8000},
xmin=250,
xmax=8000,
ylabel=Test Error,
yticklabel=\pgfmathprintnumber{\tick}\%,
]
\addplot[color=red, mark=triangle] coordinates { (250, 43.73000022172928) (500, 33.00599981546402) (1000, 26.637999832630157) (2000, 16.640000134706497) (4000, 12.907999992370605) (8000, 10.332000017166138) };
\addlegendentry{ LGA (ours) }
\addplot[color=violet, mark=square] coordinates { (250, 38.29999998211861) (500, 28.091999816894532) (1000, 21.296000277996065) (2000, 14.72800009250641) (4000, 12.063999974727631) (8000, 9.98399988412857) };
\addlegendentry{ LGA (ours) + VAT }
\addplot[color=black, mark=diamond] coordinates { (250, 51.59750033915043) (500, 43.69400017261505) (1000, 35.703999960422514) (2000, 25.665999925136568) (4000, 20.21000019311905) (8000, 14.972000074386596) };
\addlegendentry{ Supervised }
\addplot[color=blue, mark=o] coordinates { (250, 51.40266457777696) (500, 37.90818832335516) (1000, 26.92858608277225) (2000, 16.045968007659255) (4000, 13.28888225749295) (8000, 11.499151388552114) };
\addlegendentry{ VAT }
\addplot[color=green, mark=pentagon] coordinates { (250, 51.69311971949196) (500, 42.454507587767566) (1000, 34.6674251325761) (2000, 21.1729488781543) (4000, 15.223343342596543) (8000, 13.046670521165531) };
\addlegendentry{ Mean Teacher }
\addplot[color=orange, mark=x] coordinates { (250, 52.950681065085085) (500, 43.90553990960578) (1000, 36.69887038314962) (2000, 23.978029356360842) (4000, 15.804004948679193) (8000, 13.046670521165531) };
\addlegendentry{ $\Pi$-Model }
\addplot[color=black!70!green, mark=star] coordinates { (250, 51.01597130263412) (500, 39.35897196784602) (1000, 31.28168304828692) (2000, 22.430261546400068) (4000, 17.061566294272318) (8000, 13.917289914268466) };
\addlegendentry{ Pseudo Label }
\addplot[name path=upper,draw=none] coordinates { (250, 44.08400575133228) (500, 33.38243578915391) (1000, 27.037519635729613) (2000, 16.66449511827518) (4000, 13.057184533338031) (8000, 10.495144151161783) };
\addplot[name path=lower,draw=none] coordinates { (250, 43.37599469212628) (500, 32.629563841774136) (1000, 26.238480029530702) (2000, 16.615505151137814) (4000, 12.75881545140318) (8000, 10.168855883170492) };
\addplot[fill=red, fill opacity=0.15] fill between[of=upper and lower];
\addplot[name path=upper,draw=none] coordinates { (250, 38.75896608156935) (500, 29.434197994932767) (1000, 21.622717401272336) (2000, 15.04821257855068) (4000, 12.250826010682237) (8000, 10.157274234585994) };
\addplot[name path=lower,draw=none] coordinates { (250, 37.841033882667865) (500, 26.749801638856297) (1000, 20.969283154719793) (2000, 14.407787606462138) (4000, 11.877173938773025) (8000, 9.810725533671146) };
\addplot[fill=violet, fill opacity=0.15] fill between[of=upper and lower];
\addplot[name path=upper,draw=none] coordinates { (250, 52.30465566057246) (500, 44.07905081256277) (1000, 36.540937245180245) (2000, 26.128540668490505) (4000, 20.537292211197478) (8000, 15.167489302375072) };
\addplot[name path=lower,draw=none] coordinates { (250, 50.8903450177284) (500, 43.30894953266733) (1000, 34.86706267566478) (2000, 25.20345918178263) (4000, 19.882708175040623) (8000, 14.77651084639812) };
\addplot[fill=black, fill opacity=0.15] fill between[of=upper and lower];
\addplot[name path=upper,draw=none] coordinates { (250, 52.95018371039038) (500, 41.10045943139926) (1000, 27.702469987752636) (2000, 16.142952173129167) (4000, 13.482353233738053) (8000, 11.692871042144567) };
\addplot[name path=lower,draw=none] coordinates { (250, 49.855145445163544) (500, 34.715917215311066) (1000, 26.15470217779186) (2000, 15.948983842189342) (4000, 13.095411281247848) (8000, 11.305431734959662) };
\addplot[fill=blue, fill opacity=0.15] fill between[of=upper and lower];
\addplot[name path=upper,draw=none] coordinates { (250, 52.08006167198215) (500, 43.71194459468701) (1000, 35.44130903755648) (2000, 21.75361048423695) (4000, 15.513425468290507) (8000, 13.240141497410626) };
\addplot[name path=lower,draw=none] coordinates { (250, 51.30617776700177) (500, 41.19707058084812) (1000, 33.89354122759572) (2000, 20.59228727207165) (4000, 14.93326121690258) (8000, 12.853199544920436) };
\addplot[fill=green, fill opacity=0.15] fill between[of=upper and lower];
\addplot[name path=upper,draw=none] coordinates { (250, 54.9821263156586) (500, 44.38946602756589) (1000, 37.76296075249765) (2000, 24.60693436783108) (4000, 16.094211413046843) (8000, 13.33687698553318) };
\addplot[name path=lower,draw=none] coordinates { (250, 50.91923581451157) (500, 43.42161379164567) (1000, 35.63478001380159) (2000, 23.349124344890605) (4000, 15.513798484311543) (8000, 12.756464056797881) };
\addplot[fill=orange, fill opacity=0.15] fill between[of=upper and lower];
\addplot[name path=upper,draw=none] coordinates { (250, 51.9833261838596) (500, 40.71326880156169) (1000, 32.15230244138985) (2000, 22.623732522645163) (4000, 17.351772758639953) (8000, 14.014274079738385) };
\addplot[name path=lower,draw=none] coordinates { (250, 50.04861642140864) (500, 38.004675134130345) (1000, 30.411063655183984) (2000, 22.236790570154973) (4000, 16.771359829904682) (8000, 13.820305748798546) };
\addplot[fill=black!70!green, fill opacity=0.15] fill between[of=upper and lower];

\end{axis}
\end{tikzpicture}
\end{subfigure}
\begin{subfigure}[b]{.49\textwidth}
\centering
\begin{tikzpicture}
\begin{axis}[
xmode=log,
xlabel=Number of Labeled Datapoints,
grid=major,
xtick=data,
xticklabels={250,500,1000,2000,4000,8000},
xmin=250,
xmax=8000,ylabel=Test Loss,
ymax=3.453877639491069,
]
\addplot[color=red, mark=triangle] coordinates { (250, 1.5409750571250915) (500, 1.3145057280063628) (1000, 0.9350586643218994) (2000, 0.564372190028429) (4000, 0.41419744244217876) (8000, 0.34169731801748277) };
\addlegendentry{ LGA (ours) }
\addplot[color=blue, mark=o] coordinates { (250, 10.000273483276368) (500, 9.424947627544404) (1000, 2.3648165543079376) (2000, 1.685870745420456) (4000, 1.357819939851761) (8000, 0.851413498699665) };
\addlegendentry{ VAT }
\addplot[color=black, mark=diamond] coordinates { (250, 10.347285959243774) (500, 10.981996963500976) (1000, 11.890019163131715) (2000, 1.881965651512146) (4000, 1.5033702015876769) (8000, 1.0219475502371789) };
\addlegendentry{ Supervised }
\addplot[name path=upper,draw=none] coordinates { (250, 1.556973115281078) (500, 1.3279756978720914) (1000, 0.9482663358324386) (2000, 0.5695100633925733) (4000, 0.41784465824693) (8000, 0.3453269223098538) };
\addplot[name path=lower,draw=none] coordinates { (250, 1.524976998969105) (500, 1.3010357581406342) (1000, 0.9218509928113603) (2000, 0.5592343166642847) (4000, 0.4105502266374275) (8000, 0.33806771372511174) };
\addplot[fill=red, fill opacity=0.15] fill between[of=upper and lower];
\addplot[name path=upper,draw=none] coordinates { (250, 10.000525649609138) (500, 12.551043397994206) (1000, 2.438246292994661) (2000, 1.7008767813937344) (4000, 1.4117019832896698) (8000, 0.8939775161386679) };
\addplot[name path=lower,draw=none] coordinates { (250, 10.000021316943599) (500, 6.2988518570946015) (1000, 2.2913868156212143) (2000, 1.6708647094471776) (4000, 1.3039378964138522) (8000, 0.808849481260662) };
\addplot[fill=blue, fill opacity=0.15] fill between[of=upper and lower];
\addplot[name path=upper,draw=none] coordinates { (250, 10.627385740863309) (500, 11.631360151639118) (1000, 12.148550657790896) (2000, 1.9249959725036767) (4000, 1.539927712401395) (8000, 1.0870143391444436) };
\addplot[name path=lower,draw=none] coordinates { (250, 10.067186177624238) (500, 10.332633775362833) (1000, 11.631487668472534) (2000, 1.838935330520615) (4000, 1.4668126907739587) (8000, 0.9568807613299142) };
\addplot[fill=black, fill opacity=0.15] fill between[of=upper and lower];

\end{axis}
\end{tikzpicture}
\end{subfigure}
\caption{Test error and loss of various methods on CIFAR-10 \cite{cifar10} as the amount of labeled data varies.
Shaded regions indicate standard deviation over five trials. X-axis is shown on a logarithmic scale.
Accuracies for VAT, Mean Teacher, $\Pi$-Model, and Pseudo Label are taken from \citet{rse},
though we obtain similar numbers for VAT with our own implementation.}
\label{fig:cifar_results}
\end{figure*}

\begin{figure*}
\centering
\begin{subfigure}[b]{.49\textwidth}
\centering
\begin{tikzpicture}
\begin{axis}[
xmode=log,
xlabel=Number of Labeled Datapoints,
grid=major,
xtick=data,
xticklabels={250,500,1000,2000,4000,8000},
xmin=250,
xmax=8000,
ylabel=Test Error,
yticklabel=\pgfmathprintnumber{\tick}\%,
]
\addplot[color=red, mark=triangle] coordinates { (250, 34.316994526137876) (500, 12.06899207541015) (1000, 7.5599260370864725) (2000, 5.822833283004693) (4000, 4.923939493329672) };
\addlegendentry{ LGA (ours) }
\addplot[color=violet, mark=square] coordinates { (250, 27.761985176112947) (500, 9.399200870570187) (1000, 6.584203860000665) (2000, 5.1037182121754565) (4000, 4.555162548945591) };
\addlegendentry{ LGA (ours) + VAT }
\addplot[color=black, mark=diamond] coordinates { (250, 32.529963104451596) (500, 16.054855754350548) (1000, 12.39551322036678) (2000, 9.033112901079196) (4000, 7.408573856988275) (8000, 5.818991827942801) };
\addlegendentry{ Supervised }
\addplot[color=blue, mark=o] coordinates { (250, 6.767052064290972) (500, 6.137010042496616) (1000, 5.630113245174616) (2000, 5.410958904109588) (4000, 4.4109826452363485) (8000, 4.315187198784457) };
\addlegendentry{ VAT }
\addplot[color=green, mark=pentagon] coordinates { (250, 11.862942475249875) (500, 7.780845658934972) (1000, 6.00004748225351) (2000, 5.410958904109588) (4000, 4.616462097291141) (8000, 4.561762541250211) };
\addlegendentry{ Mean Teacher }
\addplot[color=orange, mark=x] coordinates { (250, 26.041238337171485) (500, 11.767147028797988) (1000, 7.438403646637074) (2000, 6.356164383561644) (4000, 5.027421001400729) (8000, 4.479570760428292) };
\addlegendentry{ $\Pi$-Model }
\addplot[color=black!70!green, mark=star] coordinates { (250, 13.99992880852778) (500, 7.191828304171317) (1000, 6.232876712328768) (2000, 4.986325110989775) (4000, 4.438474870017334) };
\addlegendentry{ Pseudo Label }
\addplot[name path=upper,draw=none] coordinates { (250, 36.189699102885164) (500, 12.447635869552457) (1000, 7.750191092436499) (2000, 5.939612665543854) (4000, 5.067454635832385) };
\addplot[name path=lower,draw=none] coordinates { (250, 32.44428994939059) (500, 11.690348281267845) (1000, 7.369660981736446) (2000, 5.706053900465531) (4000, 4.7804243508269595) };
\addplot[fill=red, fill opacity=0.15] fill between[of=upper and lower];
\addplot[name path=upper,draw=none] coordinates { (250, 29.04785344349952) (500, 9.82214445861488) (1000, 6.939945646862006) (2000, 5.238608582834258) (4000, 4.593684257817939) };
\addplot[name path=lower,draw=none] coordinates { (250, 26.476116908726375) (500, 8.976257282525495) (1000, 6.228462073139323) (2000, 4.968827841516655) (4000, 4.516640840073244) };
\addplot[fill=violet, fill opacity=0.15] fill between[of=upper and lower];
\addplot[name path=upper,draw=none] coordinates { (250, 37.401626997721664) (500, 16.429323700735253) (1000, 12.874540101395617) (2000, 9.30633393115795) (4000, 7.591122016115593) (8000, 6.141702710094488) };
\addplot[name path=lower,draw=none] coordinates { (250, 27.658299211181525) (500, 15.680387807965843) (1000, 11.916486339337943) (2000, 8.759891871000443) (4000, 7.2260256978609565) (8000, 5.496280945791114) };
\addplot[fill=black, fill opacity=0.15] fill between[of=upper and lower];
\addplot[name path=upper,draw=none] coordinates { (250, 6.890339735523849) (500, 6.38358538496237) (1000, 5.753329693027229) (2000, 5.4931506849315035) (4000, 4.534270316469225) (8000, 4.397378979606376) };
\addplot[name path=lower,draw=none] coordinates { (250, 6.643764393058095) (500, 5.890434700030863) (1000, 5.506896797322003) (2000, 5.3287671232876725) (4000, 4.287694974003472) (8000, 4.232995417962538) };
\addplot[fill=blue, fill opacity=0.15] fill between[of=upper and lower];
\addplot[name path=upper,draw=none] coordinates { (250, 12.684789060088791) (500, 7.945229220578806) (1000, 6.287718715130222) (2000, 5.493150684931507) (4000, 4.7397497685240175) (8000, 4.64402554545239) };
\addplot[name path=lower,draw=none] coordinates { (250, 11.04109589041096) (500, 7.616462097291137) (1000, 5.712376249376799) (2000, 5.328767123287669) (4000, 4.493174426058264) (8000, 4.4794995370480315) };
\addplot[fill=green, fill opacity=0.15] fill between[of=upper and lower];
\addplot[name path=upper,draw=none] coordinates { (250, 28.918021889318872) (500, 12.753448398661002) (1000, 7.767099546544479) (2000, 6.56164383561644) (4000, 5.232900453455525) (8000, 4.561762541250211) };
\addplot[name path=lower,draw=none] coordinates { (250, 23.1644547850241) (500, 10.780845658934973) (1000, 7.1097077467296685) (2000, 6.150684931506849) (4000, 4.821941549345933) (8000, 4.397378979606373) };
\addplot[fill=orange, fill opacity=0.15] fill between[of=upper and lower];
\addplot[name path=upper,draw=none] coordinates { (250, 15.808219178082192) (500, 9.219201823318535) (1000, 6.986348852116521) (2000, 6.068493150684933) (4000, 4.945229220578813) };
\addplot[name path=lower,draw=none] coordinates { (250, 12.191638438973369) (500, 5.164454785024098) (1000, 5.479404572541014) (2000, 3.904157071294616) (4000, 3.9317205194558547) };
\addplot[fill=black!70!green, fill opacity=0.15] fill between[of=upper and lower];

\end{axis}
\end{tikzpicture}
\end{subfigure}
\begin{subfigure}[b]{.49\textwidth}
\centering
\begin{tikzpicture}
\begin{axis}[
xmode=log,
xlabel=Number of Labeled Datapoints,
grid=major,
xtick=data,
xticklabels={250,500,1000,2000,4000,8000},
xmin=250,
xmax=8000,ylabel=Test Loss,
ymax=3.453877639491069,
]
\addplot[color=red, mark=triangle] coordinates { (250, 1.0088720311855492) (500, 0.48728738565068885) (1000, 0.2776279199821502) (2000, 0.22407393981579554) (4000, 0.19795730798207586) (8000, 0.1739860043364114) };
\addlegendentry{ LGA (ours) }
\addplot[color=blue, mark=o] coordinates { (250, 0.6284627677759438) (500, 0.5740112432364842) (1000, 0.48370192803187173) (2000, 0.4272120594227321) (4000, 0.3668006471669062) (8000, 0.34079562597078433) };
\addlegendentry{ VAT }
\addplot[color=black, mark=diamond] coordinates { (250, 5.514067756487647) (500, 0.9423527634214353) (1000, 0.650059277841337) (2000, 0.45453421306609193) (4000, 0.3757266727978216) (8000, 0.3326413624887037) };
\addlegendentry{ Supervised }
\addplot[name path=upper,draw=none] coordinates { (250, 1.0643220447866304) (500, 0.5249484538953099) (1000, 0.28722558221274147) (2000, 0.23962956442846034) (4000, 0.20544746169870934) (8000, 0.18112859843103107) };
\addplot[name path=lower,draw=none] coordinates { (250, 0.9534220175844681) (500, 0.44962631740606784) (1000, 0.2680302577515589) (2000, 0.20851831520313074) (4000, 0.1904671542654424) (8000, 0.16684341024179172) };
\addplot[fill=red, fill opacity=0.15] fill between[of=upper and lower];
\addplot[name path=upper,draw=none] coordinates { (250, 0.6602722979331677) (500, 0.5807708247074543) (1000, 0.5131468344075271) (2000, 0.437194509670946) (4000, 0.3792920127623899) (8000, 0.3507575477818286) };
\addplot[name path=lower,draw=none] coordinates { (250, 0.5966532376187199) (500, 0.5672516617655141) (1000, 0.45425702165621634) (2000, 0.41722960917451823) (4000, 0.3543092815714225) (8000, 0.33083370415974006) };
\addplot[fill=blue, fill opacity=0.15] fill between[of=upper and lower];
\addplot[name path=upper,draw=none] coordinates { (250, 9.272579083311488) (500, 0.9747545193941732) (1000, 0.6689488702430827) (2000, 0.46329861869582667) (4000, 0.3825730325281273) (8000, 0.3446130543882808) };
\addplot[name path=lower,draw=none] coordinates { (250, 1.7555564296638058) (500, 0.9099510074486973) (1000, 0.6311696854395913) (2000, 0.4457698074363572) (4000, 0.36888031306751595) (8000, 0.32066967058912654) };
\addplot[fill=black, fill opacity=0.15] fill between[of=upper and lower];

\end{axis}
\end{tikzpicture}
\end{subfigure}
\caption{Test error and loss of various methods on SVHN \cite{svhn} as the amount of labeled data varies.
Shaded regions indicate standard deviation over five trials. X-axis is shown on a logarithmic scale.
Accuracies for VAT, Mean Teacher, $\Pi$-Model, and Pseudo Label are taken from \citet{rse},
though we obtain similar numbers for VAT with our own implementation.}
\label{fig:svhn_results}
\end{figure*}
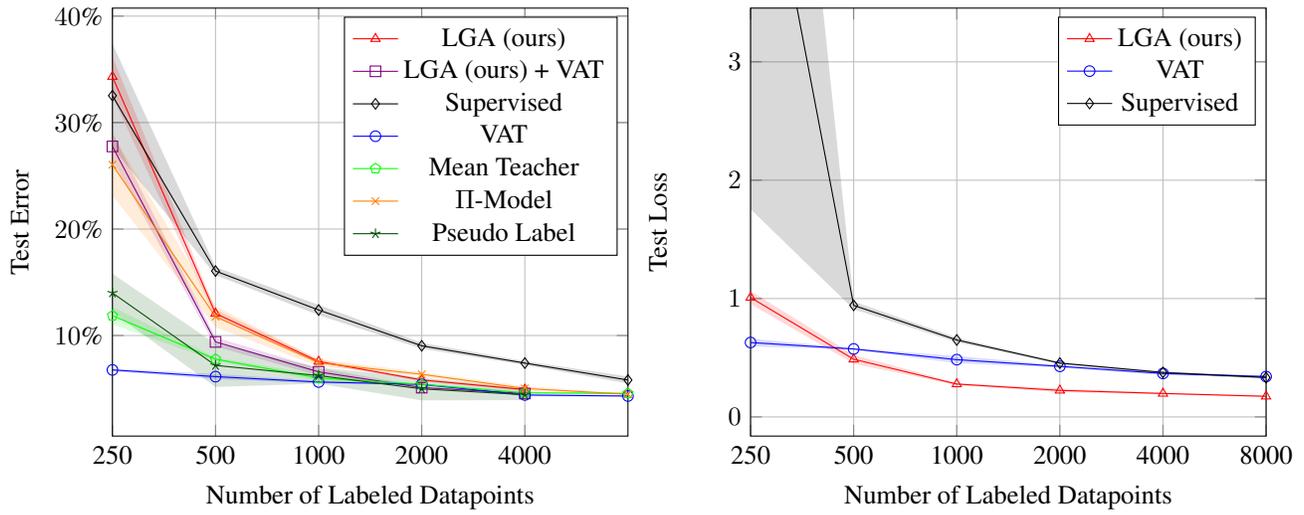

\begin{figure*}
\centering
\begin{subfigure}[b]{.49\textwidth}
\centering
\begin{tikzpicture}
\begin{axis}[
xmode=log,
xlabel=Number of Labeled Datapoints,
grid=major,
xtick=data,
xticklabels={250,500,1000,2000,4000,8000},
xmin=250,
xmax=8000,
ylabel=Test Error,
yticklabel=\pgfmathprintnumber{\tick}\%,
]
\addplot[color=red, mark=triangle] coordinates { (250, 21.48000032901764) (500, 16.476000237464905) (1000, 14.220000147819519) (2000, 11.119999933242799) (4000, 9.555999755859375) (8000, 8.695999956130981) };
\addlegendentry{ LGA (ours) }
\addplot[color=blue, mark=o] coordinates { (250, 21.916000270843504) (500, 19.24800009727478) (1000, 15.539999961853027) (2000, 12.471999859809875) (4000, 10.255999970436097) (8000, 8.73599989414215) };
\addlegendentry{ VAT }
\addplot[color=violet, mark=square] coordinates { (250, 20.31600022315979) (500, 16.37600016593933) (1000, 13.936000084877014) (2000, 10.975999665260314) (4000, 9.639999890327454) (8000, 8.483999919891357) };
\addlegendentry{ LGA (ours) + VAT }
\addplot[color=black, mark=diamond] coordinates { (250, 23.316000270843507) (500, 19.560000205039977) (1000, 16.108000183105467) (2000, 12.824000072479247) (4000, 10.895999789237976) (8000, 9.059999775886535) };
\addlegendentry{ Supervised }
\addplot[name path=upper,draw=none] coordinates { (250, 21.780399992101) (500, 16.6326654726344) (1000, 14.345698254294321) (2000, 11.305472465635862) (4000, 9.689506269832032) (8000, 8.88841621664734) };
\addplot[name path=lower,draw=none] coordinates { (250, 21.179600665934277) (500, 16.31933500229541) (1000, 14.094302041344717) (2000, 10.934527400849735) (4000, 9.422493241886718) (8000, 8.503583695614623) };
\addplot[fill=red, fill opacity=0.15] fill between[of=upper and lower];
\addplot[name path=upper,draw=none] coordinates { (250, 22.47132346598904) (500, 19.6979958259635) (1000, 15.837993331016198) (2000, 12.67658715269761) (4000, 10.65306940260706) (8000, 9.181672421532895) };
\addplot[name path=lower,draw=none] coordinates { (250, 21.360677075697968) (500, 18.79800436858606) (1000, 15.242006592689856) (2000, 12.26741256692214) (4000, 9.858930538265135) (8000, 8.290327366751406) };
\addplot[fill=blue, fill opacity=0.15] fill between[of=upper and lower];
\addplot[name path=upper,draw=none] coordinates { (250, 20.679406282575176) (500, 16.864164132220967) (1000, 14.152296192851326) (2000, 11.252376153095275) (4000, 9.845133985461473) (8000, 8.718742278266252) };
\addplot[name path=lower,draw=none] coordinates { (250, 19.952594163744404) (500, 15.887836199657695) (1000, 13.719703976902702) (2000, 10.699623177425353) (4000, 9.434865795193435) (8000, 8.249257561516462) };
\addplot[fill=violet, fill opacity=0.15] fill between[of=upper and lower];
\addplot[name path=upper,draw=none] coordinates { (250, 23.50250502442777) (500, 20.002357511191086) (1000, 16.371697015084106) (2000, 13.01807209249182) (4000, 11.092122218159092) (8000, 9.324725901620587) };
\addplot[name path=lower,draw=none] coordinates { (250, 23.129495517259244) (500, 19.117642898888867) (1000, 15.84430335112683) (2000, 12.629928052466674) (4000, 10.69987736031686) (8000, 8.795273650152483) };
\addplot[fill=black, fill opacity=0.15] fill between[of=upper and lower];

\end{axis}
\end{tikzpicture}
\end{subfigure}
\begin{subfigure}[b]{.49\textwidth}
\centering
\begin{tikzpicture}
\begin{axis}[
xmode=log,
xlabel=Number of Labeled Datapoints,
grid=major,
xtick=data,
xticklabels={250,500,1000,2000,4000,8000},
xmin=250,
xmax=8000,ylabel=Test Loss,
ymax=3.453877639491069,
]
\addplot[color=red, mark=triangle] coordinates { (250, 0.7602726247310639) (500, 0.6445479638576508) (1000, 0.48637902683019635) (2000, 0.39167284023761745) (4000, 0.35256138160824774) (8000, 0.3078018850833178) };
\addlegendentry{ LGA (ours) }
\addplot[color=blue, mark=o] coordinates { (250, 5.6850322661399835) (500, 1.5050006489753724) (1000, 1.1749089950323106) (2000, 0.8716301952302457) (4000, 0.7472274166345596) (8000, 0.5041134163141251) };
\addlegendentry{ VAT }
\addplot[color=black, mark=diamond] coordinates { (250, 1.8429118769168853) (500, 1.432987787723541) (1000, 1.1223439317196606) (2000, 0.8196064738035203) (4000, 0.709927985906601) (8000, 0.6145550852268934) };
\addlegendentry{ Supervised }
\addplot[name path=upper,draw=none] coordinates { (250, 0.7868040646238907) (500, 0.66282580739973) (1000, 0.5088718143038038) (2000, 0.3957409531483249) (4000, 0.3639264390484377) (8000, 0.32065639027969806) };
\addplot[name path=lower,draw=none] coordinates { (250, 0.7337411848382371) (500, 0.6262701203155716) (1000, 0.4638862393565889) (2000, 0.38760472732691) (4000, 0.3411963241680578) (8000, 0.29494737988693753) };
\addplot[fill=red, fill opacity=0.15] fill between[of=upper and lower];
\addplot[name path=upper,draw=none] coordinates { (250, 10.018270281501877) (500, 1.5541175738056532) (1000, 1.2049297749240546) (2000, 0.9054547188626786) (4000, 0.7803575867802757) (8000, 0.5300229034880063) };
\addplot[name path=lower,draw=none] coordinates { (250, 1.3517942507780898) (500, 1.4558837241450917) (1000, 1.1448882151405666) (2000, 0.8378056715978127) (4000, 0.7140972464888435) (8000, 0.47820392914024396) };
\addplot[fill=blue, fill opacity=0.15] fill between[of=upper and lower];
\addplot[name path=upper,draw=none] coordinates { (250, 1.8852183463266357) (500, 1.4653690991871895) (1000, 1.1380603117271972) (2000, 0.8292344593296965) (4000, 0.7297129889664351) (8000, 0.6238013888203163) };
\addplot[name path=lower,draw=none] coordinates { (250, 1.800605407507135) (500, 1.4006064762598927) (1000, 1.106627551712124) (2000, 0.809978488277344) (4000, 0.6901429828467669) (8000, 0.6053087816334705) };
\addplot[fill=black, fill opacity=0.15] fill between[of=upper and lower];

\end{axis}
\end{tikzpicture}
\end{subfigure}
\caption{Test error and loss of various methods on Fashion MNIST \cite{xiao2017/online}
as the amount of labeled data varies.
Shaded regions indicate standard deviation over five trials.
X-axis is shown on a logarithmic scale.
Hyperparameters for each method were transferred from CIFAR-10 and SVHN
by trying both sets of hyperparameters and taking the best.}
\label{fig:fashion_mnist_results}
\end{figure*}
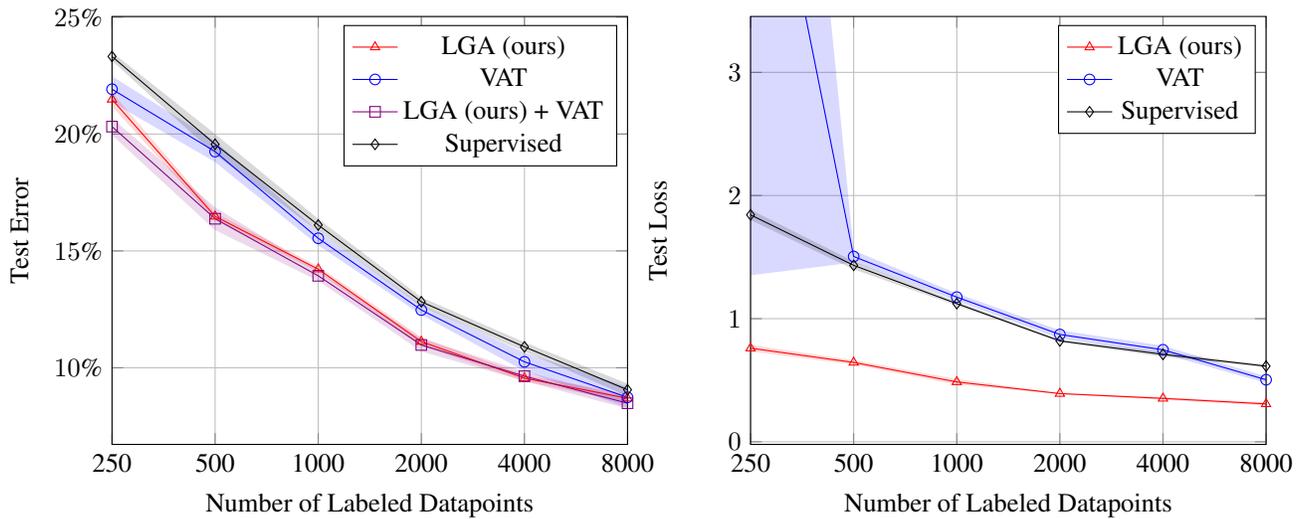

Our main results are shown in
Figures~\ref{fig:cifar_results}-\ref{fig:fashion_mnist_results}
and Table~\ref{table:many_results}.
\Om{} achieves state-of-the-art accuracy on CIFAR-10 with 4000 labels
(Table~\ref{table:many_results}).
We compare our numbers to those reported in \citet{rse},
rather than those originally reported in the literature,
because this allows us to use the same architecture
and labeled/unlabeled splits for all comparisons.
\citet{rse} showed that a fully supervised model with
Shake-Shake regularization \cite{Gastaldi2017ShakeShakeR}
is competitive with the best SSL methods using the WRN-28-2 architecture,
highlighting the importance of comparing SSL techniques using a standardized architecture.
We refer to \citet{rse} for a comparison with the numbers from the published literature.
\citet{manifold_mixup} use WRN-28-2, but we do not include their results in
Table~\ref{table:many_results} because they augment the model
with additional regularization such as dropout \cite{dropout}.

\Om{} consistently achieves better test loss than other SSL methods
(Figures~\ref{fig:cifar_results}-\ref{fig:fashion_mnist_results}),
even when its accuracy is lower than other SSL methods (Figure~\ref{fig:svhn_results}).
This indicates that \om{} is better at avoiding highly confident incorrect predictions,
which makes it well-suited for applications where
measuring model uncertainty is important.
Test loss is not reported in \citet{rse},
so we used our reimplementation of VAT for loss comparisons,
which we found to achieve similar accuracy to the implementation of \citet{rse}.

\begin{table}
\caption{
Test error rates of SSL methods on CIFAR-10 and SVHN. \\
$\mu \pm \sigma$ indicates mean $\mu$ and standard deviation $\sigma$ over five trials.
}
\label{table:many_results}
\begin{tabular}{lrr}
\toprule
Method & CIFAR-10 & SVHN \\
& 4000 labels & 1000 labels \\
\midrule
LGA (ours) & 12.91 $\pm$ .15 & 7.56 $\pm$ .19 \\
LGA (ours) + VAT\textsuperscript{1} & \textbf{12.06 $\pm$ .19} & 6.58 $\pm$ .36 \\
VAT\textsuperscript{1} & 13.86 $\pm$ .27 & 5.63 $\pm$ .20 \\
VAT\textsuperscript{1} + EntMin\textsuperscript{2} & 13.13 $\pm$ .39 & \textbf{5.35 $\pm$ .19} \\
$\Pi$-Model\textsuperscript{3}  & 16.37 $\pm$ .63 & 7.19 $\pm$ .27 \\
Mean Teacher\textsuperscript{4}  & 15.87 $\pm$ .28 & 5.65 $\pm$ .47 \\
Pseudo-Label\textsuperscript{5} & 17.78 $\pm$ .57 & 7.62 $\pm$ .29 \\
Supervised & 20.26 $\pm$ .38 & 12.83 $\pm$ .47 \\
\bottomrule
\end{tabular}

\textsuperscript{1} \citet{vat} \\
\textsuperscript{2} \citet{entmin} \\
\textsuperscript{3} \citet{pimodel,pimodel2} \\
\textsuperscript{4} \citet{meanteacher} \\
\textsuperscript{5} \citet{Lee2013PseudoLabelT}
\end{table}

\citet{rse} found that VAT achieved the best performance of the methods they evaluated, so we evaluate \om{} combined with VAT.
We find that in many cases,
the combination achieves better performance than either method individually
(Figure~\ref{fig:cifar_results} and Figure~\ref{fig:fashion_mnist_results}).

\section{Related Work}
\label{sec:related_work}
\subsection{Graph-Based Methods}
Label propagation was first introduced by \citet{Zhu02learningfrom}.
It operates on a weighted graph whose vertices are the input points,
and whose edge weights represent the similarity between the points connected by the edge.
This graph can be provided as part of the input, or formed from unstructured data by creating
an edge for each pair of points whose weight is given by an appropriate similarity metric.
The label propagation algorithm learns labels for each vertex by
repeatedly setting each unlabeled point's label to the weighted average of its neighbors' labels.
This process causes the labels, originating from the labeled points, to diffuse through the graph.
It converges to a solution where each node's label is the weighted average of its neighbors,
which is appealing because it reflects the semi-supervised smoothness assumption that
close points in high-density regions should have similar labels.

However, when dealing with complex, unstructured data such as images, it is difficult to
define a similarity metric that reflects meaningful variations in the input space.
\citet{Husser2017LearningBA} addressed this issue by
learning a network to map the input data into a latent
space where the dot product metric was meaningful.
The network was trained using the objective of \textit{cycle-consistency}:
a random walk in the latent space
beginning at a point in class $c$ should end at a point in class $c$.
\citet{pmlr-v80-kamnitsas18a} also learned latent space embeddings;
their embeddings were learned by using label propagation within training batches.
One attractive quality of these methods is that the embeddings are trained to be sensitive to 
variations in the input $x$ which affect the output $y$; that is, they represent information
useful for modeling $p(y \, | \, x)$, not $p(x)$.
Ladder Networks \cite{Rasmus2015SemisupervisedLW} were introduced because of this issue, 
specifically,
that autoencoders are too sensitive to variations in $p(x)$
to provide a good auxiliary task for semi-supervised learning.
Since our method uses the gradient of a classifier trained to predict $p(y\,|\,x)$,
we argue that it, too, discards information that is relevant to modeling $p(x)$ but irrelevant for
modeling $p(y\,|\,x)$.

\subsection{Consistency Regularization}
The current state-of-the-art methods for semi-supervised image classification are
consistency regularization methods.
Consistency regularization defines a transformation to the input data that should not affect the
classification, then regularizes the model so that its output is invariant to this transformation
on the unlabeled data.
The $\Pi$-Model \cite{pimodel,pimodel2} regularizes the model to give the same prediction when run twice on the same point.
Weights averaging methods, which include Mean Teacher \cite{meanteacher} and
fast-SWA \cite{athiwaratkun2018there},
extend the $\Pi$-Model by additionally using an exponential moving average of the model weights
for the target predictions.
Virtual adversarial training \cite{vat} uses multiple passes through the network to compute
adversarial perturbations,
which approximately maximize the difference in classification loss subject to
an $L_2$ norm constraint.
Consistency regularization techniques are orthogonal to our goal of obtaining meaningful
similarity metrics between points in the training data,
and we find in Section~\ref{sec:evaluation} that virtual adversarial training
can be successfully combined with \om{}.

\subsection{Gradient Similarity}
Recent work has explored the use of gradient similarity as a way of using a deep model's
latent representations to create semantically meaningful metrics.
\citet{Koh2017UnderstandingBP} used influence functions
to construct adversarial attacks and determine the influence
of individual training points on the model output.
Influence functions generalize the gradient dot product, and are equivalent in the
case that the Hessian is the identity matrix.
\citet{sim_rl} used gradient similarity to determine which auxiliary tasks are helpful
for learning a task in reinforcement learning,
and \citet{sim_adv} used gradient similarity to detect adversarial examples.

\section{Conclusion}
We have presented a novel algorithm for semi-supervised learning by imputing labels.
Through analyzing simple models and datasets, we gave intuition for why \om{} works.
We evaluated \om{} on standard benchmark datasets and demonstrated performance
comparable to or exceeding that of state-of-the-art consistency regularization methods.
We consider this a promising result for semi-supervised learning techniques
that use gradient similarity.

\bibliography{ms}
\bibliographystyle{icml2019}

\ifdefined\arxiv
\onecolumn
\appendix
\newcommand{\sg}[1]{{\text{StopGradient}(#1)}}

\section{Proof of Proposition 1}

\begingroup
\def\thetheorem{1}
\begin{proposition}
Let $i \ne j$. Then the value of $c_{k,i}$ does not depend on
$\lambda^\ell_j$ or $\lambda^u_j$.
\end{proposition}
\addtocounter{theorem}{-1}
\endgroup

\begin{proof}
Let $b = \frac{1}{n_\ell} X^\top_\ell y_\ell$.
\begin{align}
g^\ell_k &= \frac{1}{n_\ell} X^\top_\ell(X_\ell \theta_k - y_\ell) \\
g^u_k &= \frac{1}{n_u} X^\top_u(X_u \theta_k - y^u_k)
\end{align}
\begin{align}
\Vert g_\ell - g_u \Vert^2
&= \left\Vert
\frac{1}{n_u} X^\top_u y^u_k
- \frac{1}{n_\ell} X_\ell^\top y_\ell
+ \Big(
  \frac{1}{n_\ell} X^\top_\ell X_\ell
- \frac{1}{n_u} X^\top_u X_u
\Big) \theta_k
\right\Vert^2 \\
&= \left\Vert
\frac{1}{n_u} X^\top_u y^u_k
- b
+ \Big(
  \frac{1}{n_\ell} X^\top_\ell X_\ell
- \frac{1}{n_u} X^\top_u X_u
\Big) \theta_k
\right\Vert^2 \\
&= \sum_{i=1}^m \left[ e_i^\top \left(
\frac{1}{n_u} X^\top_u y^u_k
- b
+ \Big(
  \frac{1}{n_\ell} X^\top_\ell X_\ell
- \frac{1}{n_u} X^\top_u X_u
\Big) \theta_k
\right) \right]^2 \\
\label{eq:sum}
&= \sum_{i=1}^m \left[ e_i^\top
\left( \frac{1}{n_u} X^\top_u y^u_k - b \right)
+ (\lambda^\ell_i - \lambda^u_i) \theta_{k,i}
\right]^2
\end{align}
Let $r_i$ denote the expression in square brackets in Equation~\ref{eq:sum}.
We define StopGradient as a function that returns its argument but whose derivative is zero
everywhere.
\begin{align}
\Vert g_\ell - g_u \Vert^2
&= \sum_{i=1}^m r_i^2 \\
\Vert g_\ell - g_u \Vert^2_\text{normalized}
&= \sum_{i=1}^m \frac{r_i^2}{\sg{r_i^2}} \\
\nabla_{y^u_k} \frac{1}{2} \Vert g_\ell - g_u \Vert^2_\text{normalized}
&= \sum_{i=1}^m \frac{\nabla_{y^u_k} \, r_i}{r_i} \\
&= \sum_{i=1}^m \frac{1}{r_i}
\nabla_{y^u_k} \left( \frac{1}{n_u} e_i^\top X^\top_u y^u_k \right) \\
\label{eq:linear_update}
&= \frac{1}{n_u} X_u \sum_{i=1}^m \frac{1}{r_i}
e_i
\end{align}

This allows us to show that for all $k$, there exists $a_k \in \mathbb{R}^m$ such that
$y^u_k = \frac{1}{n_u} X_u a_k$.
The proof is by induction on $k$.
In the base case, we initialize $y^u_0=0$.
In the inductive case, Equation~\ref{eq:linear_update} shows that the update to
$y^u_k$ is linear in $X_u$.

The update rule for $\theta_k$ can then be written as:
\begin{align}
\theta_{k+1}
&= \theta_k - \alpha_\theta g^u_k \\
\label{eq:theta_update}
&= \theta_k + \alpha_\theta \left( a_k - \frac{1}{n_u} X^\top_u X_u \theta_k \right)
\end{align}

We are ready to prove the proposition.
Let $i \ne j$.
By induction on $k$, we will show that $\theta_{k,i}$ and $a_{k,i}$ do not depend on
$\lambda^\ell_j$ or $\lambda^u_j$.
In the base case, the initialization is $\theta_0=a_0=0$.
In the inductive case, by Equation~\ref{eq:theta_update},
the property holds for $\theta_{k+1,i}$ provided that it holds
for $\theta_{k,i}$ and $a_{k,i}$.
Considering $a_k$, note that $a_{k+1,i}-a_{k,i}=1/r_i$,
and from Equation~\ref{eq:sum} we see that $r_i$ does not depend on
$\lambda^\ell_j$ or $\lambda^u_j$.

Having shown that $\theta_{k,i}$ does not depend on $\lambda^\ell_j$ or $\lambda^u_j$,
we complete the proof by writing $c_{k,i}$ in terms of $\theta_{k,i}$:
\begin{equation}
c_{k,i} = 
\frac{\lambda^\ell_i}
{e_i^\top ( \frac{1}{n_\ell} X^\top_\ell y_\ell ) }
\theta_{k,i} 
\end{equation}
\end{proof}
\fi

\end{document}